\documentclass[letterpaper, 10 pt, conference]{ieeeconf}   
\IEEEoverridecommandlockouts 
\usepackage{graphicx}
\usepackage{amssymb}
\usepackage{amsmath}
\usepackage{multirow}
\usepackage{booktabs}
\usepackage{threeparttable}
\makeatletter
\let\NAT@parse\undefined

\newcommand{\Rmnum}[1]{\expandafter\@slowromancap\romannumeral #1@}
\makeatother
\usepackage[colorlinks, linkcolor=red, anchorcolor=blue, citecolor=green]{hyperref}

\title
{\LARGE \bf
Hierarchical Reinforcement Learning Framework towards \\Multi-agent Navigation
}

\author{Wenhao Ding$^{1}$, Shuaijun Li$^{2}$ and Huihuan Qian$^{2}$
\thanks{This research is supported by the NSFC project U1613226 from the State Joint Engineering Lab and Shenzhen Engineering Lab on Robotics and Intelligent Manufacturing, China.}
\thanks{$^{1}$Wenhao Ding is with the Department of Electronic Engineering, Tsinghua University, Haidian District, Beijing, China {\tt\small dingwenhao95@gmail.com}}%
\thanks{$^{2}$Shuaijun Li and Huihuan Qian are with the Robotics and Artificial Intelligence Laboratory, The Chinese University of Hong Kong, Shenzhen, Shenzhen, China {\tt\small \{sjli01, hhqian\}@mae.cuhk.edu.hk}}%
}

\begin{document}
\maketitle

\begin{abstract}

In this paper, we propose a navigation algorithm oriented to multi-agent environment. This algorithm is expressed as a hierarchical framework that contains a \textit{Hidden Markov Model (HMM)} and a \textit{Deep Reinforcement Learning (DRL)} structure. For simplification, we term our method \textit{Hierarchical Navigation Reinforcement Network (HNRN)}. In high-level architecture, we train an HMM to evaluate the agent’s perception to obtain a score. According to this score, adaptive control action will be chosen. While in low-level architecture, two sub-systems are introduced, one is a differential target-driven system, which aims at heading to the target; the other is a collision avoidance \textit{DRL} system, which is used for avoiding dynamic obstacles. The advantage of this hierarchical structure is decoupling the target-driven and collision avoidance tasks, leading to a faster and more stable model to be trained. The experiments indicate that our algorithm has higher learning efficiency and rate of success than traditional \textit{Velocity Obstacle (VO)} algorithms or hybrid \textit{DRL} method.

\end{abstract}

\section{INTRODUCTION}

The navigation problem of mobile robots has always been a hot topic. At present, researches on static obstacle scenes has made great progress, while there is still huge space for dynamic obstacle avoidance tasks. Moreover, dynamic collision avoidance navigation has great application prospects, such as multi-vehicle interaction scenarios in automatic driving and collaborative scenes in logistic warehouses.

Most of the dynamic navigation scenarios can be simplified into such an environment. After knowing the location of their target points, agents need to use their own onboard sensors (2D laser in this paper) to perceive the environment. They utilize the result of perception to avoid the dynamic obstacles (possibly other agents) in the process of moving forward and finally reach the destination quickly and safely. There have been many ways to solve such a hypothetical situation.

First of all, we think of globally centralized navigation algorithms. In this kind of algorithm, it is assumed that a central node controls the current location and the target position of all agents. In this way, we can avoid collision through some pre-planning paths, achieving the globally optimal route. But in practice, most scenarios do not have such a central node that collecting all participants' positions and targets. Moreover, even surrounding agents can only be perceived indirectly and imperfectly through real-world sensors. Based on the above limitations, more algorithms focus on the decentralized navigation.

For dynamic collision avoidance scenarios, there is a class of \textit{VO} algorithms and their variants that can achieve stunning results. Such algorithms use geometric relations to calculate the possible path of safe distance in real time. Experiments manifest that these methods could be effectively implemented in swarm navigation in large-scale situations. In these works, the authors assume that each agent knows the exact location and speed of surrounding agents, which in fact omits the precision environmental perception. Besides, a large number of hyper-parameters and prior assumptions are prerequisites.

\begin{figure}[]
\centering
\includegraphics[width=8cm]{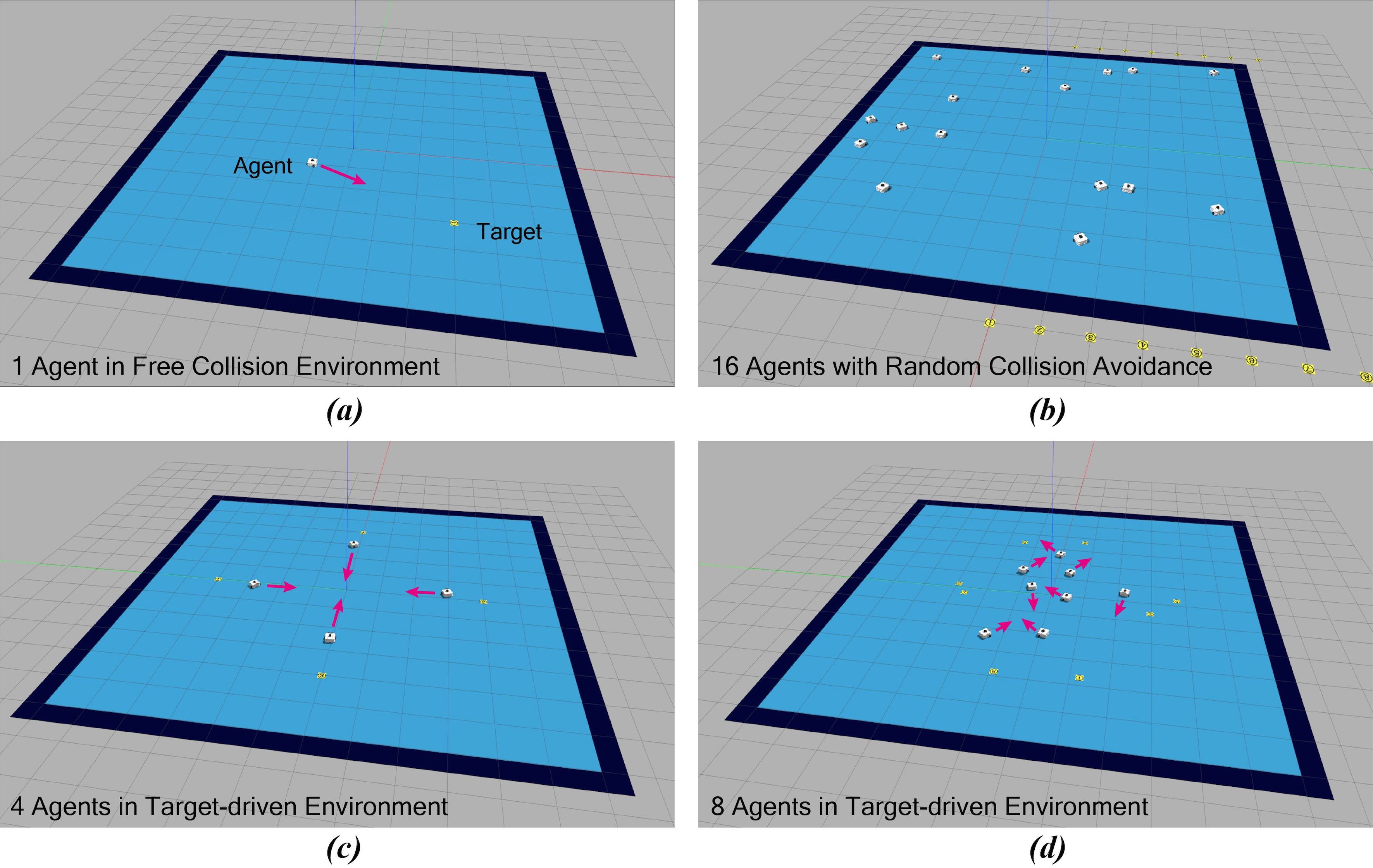}
\caption{\textbf{Schematic Diagram of Hierarchical Navigation Reinforcement Network.} (a) represents the target-driven stage, (b) represents the collision avoidance stage.}
\label{fig1}
\end{figure} 

In order to keep away from the shortcomings of above traditional methods, researchers attempts to use end-to-end \textit{DRL} framework for perception and decision making with the development of \textit{DNN} and \textit{DRL} tools in the field of robotics. The core of \textit{DRL} methods is to set up a reward function to guide the agent. When the actions taken by the agent are beneficial to the completion of the task, a high reward is given. On the contrary, when the agent makes some mistakes, it will be punished through receiving a low reward. Common problems of \textit{DRL} methods are the difficulty of training and the slow speed of convergence. Especially in dynamic navigation scenarios, the whole framework not only completes the perception and feature extraction of front-end but also contains a complex reward function for the task of target-driven and collision avoidance. Research shows that such a multitasking complex reward often makes the reinforcement network more difficult to train and easier to fall into local minimum value.

Inspired by some recent works of \textit{DRL}, we propose a hierarchical reinforcement learning method called \textit{HNRN} for dynamic navigation scenarios and the schematic diagram is shown in Figure~\ref{fig1}. In high-level structure, we build an HMM named \textit{Evaluation Net} to classify the environment perception. Then this \textit{HMM} gives a hazard coefficient, which is used to control the low-level decision-making process. The structure of low-level is divided into two parts, corresponding to two sub-tasks of dynamic navigation: target-driven and collision avoidance. 

In target-driven sub-task, we use the position and the angle difference between the target point and the current point to drive the agent, which is the best decision for agents in the absence of obstacles. While in the collision avoidance part, we exploit the \textit{DRL} framework to train a model that accomplishes the dynamic obstacle avoidance task. Its reward function involves only a single task, greatly reducing the training difficulty of the model.

According to the experience of human beings, we move forward to the target directly in an open environment, but at congested traffic junctions or in the middle of the crowd, the prior task is to find a relatively low complexity path, sometimes even the opposite direction to the target. Once entering a relatively simple scene, it is much easier to move towards the target efficiently. In the experiment section, we prove that this hierarchical framework greatly reduces the time of training and keeps the same success rate with other algorithms.

\section{RELATED WORKS}

In this section, we will introduce some works and technology related to our methods, discuss their insight and some limitations, and then summarize the contribution of this article.

\subsection{Velocity Obstacle Algorithms} 
To tackle the problem of dynamic collision avoidance, one of the most important kinds of algorithm is \textit{VO} algorithm. The fundamental thought of it is to preclude velocities that possibly lead to collision. The original work of \textit{VO} \cite{8} helps the agent simply avoid possible collisions in the future and has achieved great results. On this basis, \textit{RVO} \cite{9} solves the tremor problem in \cite{8}. It assumes that other agents will also avoid obstacles instead of remaining stationary. \textit{OCRA} \cite{10} further extends \textit{RVO} \cite{9} to more robots and more complex condition by optimizing the original algorithm. 

After the emergence of the \textit{OCRA}, a series of deformation algorithms appear later, such as \textit{HRVO} \cite{11}, \textit{NH-OCRA} \cite{16}, \textit{OCRA-DD} \cite{17}. Although the effectiveness of these traditional algorithms is outstanding, they also have some limitations. First, the premise of this kind of algorithm for \textit{VO} is that each agent has a sensor of perfect, which can collect the position, speed, and shape of all surrounding agents in the environment. Of course, this is usually not implementable in the actual scene. Secondly, as a non-learning-based algorithm, there are many hyper-parameters to be adjusted, and the process of adjusting parameters is often time-consuming and laborious.

\subsection{Navigation with Learning-based Methods}

In addition to the above traditional \textit{VO} algorithm, another part of the attention is focused on learning-based methods. First of all, a relatively simple way is to use supervised learning \cite{30, 31, 32, 5}. Through a large number of manually calibrated data or traditional behavioral data, obstacle avoidance navigation problem in the same scenarios could be solved. \cite{30} and \cite{31} directly use \textit{CNN} to process raw RGB image and depth image respectively. \cite{32} constructs an autoencoder to encode and decode the perception and output control actions. \cite{33} introduces imitation learning to solve this problem.

Although the convergence speed of these methods is fast, supervised learning methods have disadvantages of poor transferability and require a larger amount of annotation data.

To compensate for these shortcomings, researchers begin to utilize the idea of \textit{DRL}. The rise of \textit{DRL} comes from an article of \textit{Deepmind} \cite{33}. They propose \textit{Deep Q Network (DQN)} that introduces deep representation learning into reinforcement learning and exceeds human players in Atari games. Then a series of continuous action learning algorithms appears, such as \textit{ Deep Deterministic Policy Gradient (DDPG)} \cite{34}, \textit{Trust Region Policy Optimization (TRPO)} \cite{35} and \textit{Proximal Policy Optimization (PPO)} \cite{36}. These algorithms are widely used in many fields, including mobile robot navigation. 

Combined with the framework of \textit{DDPG}, \cite{4} trained a model that navigates in a static environment with 10-dimensional laser range findings as input. Chen \emph{et al.} propose decentralized non-communicating algorithms \cite{19, 20, 21} for multi-agent obstacle avoidance navigation tasks in a dynamic environment. \cite{6, 7} focus on the generation of social behavior in the dynamic scenarios through \textit{Generative Adversarial Networks (GAN)} or \textit{Generative Adversarial Imitation Learning (GAIL)} algorithm. In addition, \cite{2} proposes a \textit{PPO} algorithm with multiple robots and accelerated the training speed of the enhanced learning model by pre-training in a simple environment.

As known to us all, the biggest limitation of \textit{DRL} is the slow and unstable training process, especially for the problem of dynamic obstacle avoidance navigation that requires the learning of obstacle avoidance and target-driven tasks at the same time. For example, \cite{2} and \cite{3} train their models by designing a very complex reward function. While \cite{37} believes that it is easier for complex reward to make the agent fall into the local minimum.

\begin{figure*}[]
\centering
\includegraphics[width=17cm]{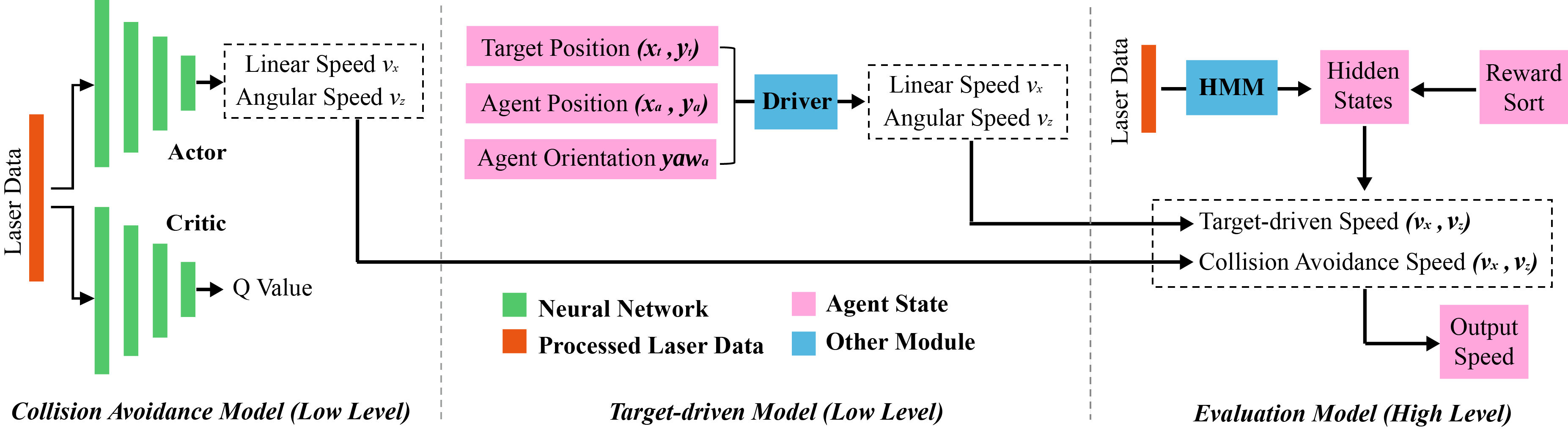}
\caption{\textbf{System Structure of Hierarchical Navigation Reinforcement Network.} The whole system is divided into three blocks. The first two blocks are low-level model that output candidate speeds. The first model has a \textit{Actor-Critic} structure and the second one has a differential target driven model. The last block contains the high-level module.}
\label{fig2}
\end{figure*}

\subsection{Hierarchical Task and Motion Planning}

Aiming at solving the motion planning problem, a kind of hierarchical method has achieved great results. \cite{38} and \cite{39} combine task and motion planning. Firstly, they judge the type of the task and then select the specific control strategy according to the current task. Hierarchical ideas are also used in the \textit{DRL} framework. \cite{25} proposes a cascaded \textit{DRL} framework to determine the current perception type at the high level and select specific behavior in the low level, according to a kind of pre-trained strategy. Ma \emph{et al.}\cite{42} also introduce a hierarchical framework for coordinating task-level and motion-level operations.

Hierarchical structures, along with many advantages, are quite similar to human decision-making process. Most significantly, decoupling task type and the specific decision-making process greatly speeds up the training stage and reduce training difficulty. In \cite{25, 26}, the high-level structures are completely independent, which is not reasonable. Thus, how to establish the connection between tasks is an open problem worthy of study. In addition, how to accurately determine current task is also an important research point. \cite{25} utilizes supervised learning, but our perception of the environment is very complex in many cases, leaving this method not applicable.

\subsection{Our contribution}

Inspired by the advantages and limitations of the above methods, we propose the \textit{HNRN} to solve navigation problems in dynamic environments. Our contributions are as follow:
\begin{itemize}
\item Using reinforcement learning method to solve the dynamic obstacle avoidance problem without a lot of training data to be annotated and hyper-parameters to be adjusted.
\item Decoupling the target-driven and the collision avoidance task to speed up the training process of \textit{DRL} greatly.
\item Using hierarchical architecture to extract high-level results to select specific strategy actions at low-level.
\item Utilizing the scheduling of HMM to reasonably classify the current task.
\end{itemize}

\begin{figure*}[]
\centering
\includegraphics[width=17.5cm]{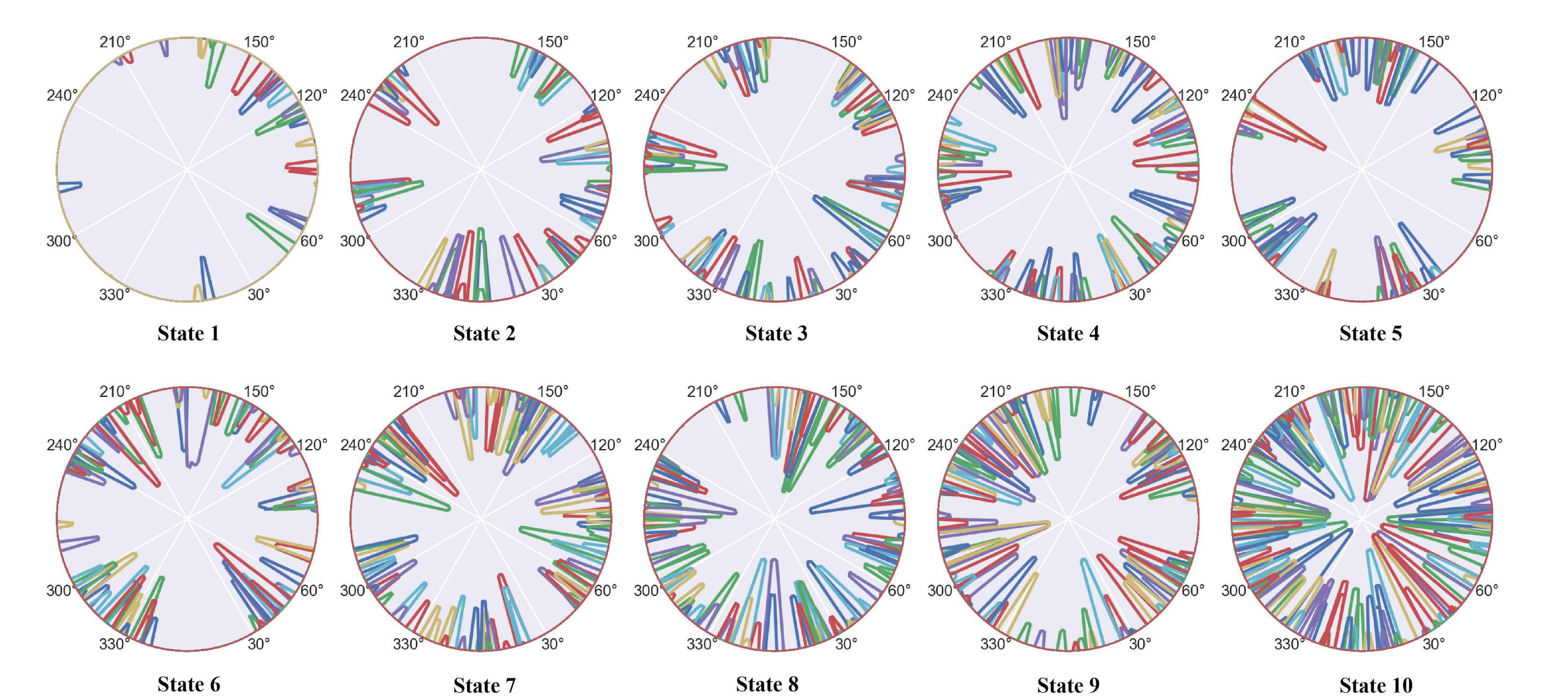}
\caption{\textbf{Visualization of HMM States’ Results.} We set the number of \textit{HMM} state to 10, and this figure shows 20 samples randomly selected for each state. From left to right, the state number become bigger, and the state becomes more complicated. For example, state 1's samples have less obstacles in laser detection, but state 10's samples almost bump into obstacles.}
\label{fig3}
\end{figure*}

\section{METHOD}

In this section, we will introduce the structure of the whole system in detail. The structural diagram of the whole system is shown in  Figure~\ref{fig2}.

\subsection{Brief Description}

First, our system is divided into two levels, the part of high level is responsible for the perception of the overall environment, and then gives a score on the current risk. According to this score, we choose the behavior of low level. Low-level system includes two solutions to different scenarios: for target driven tasks, we directly use a differential drive system; for dynamic collision avoidance tasks, we train a DRL network. Next, we will introduce each module separately.

\subsection{High-level Evaluation Module}

The middle part of Figure~\ref{fig2} is the high-level module, which we call \textit{Evaluation Module}. Based on the characteristics of the continuity of agent motion, we use \textit{HMM} to implement this module. \cite{40} and \cite{41} have also proved the competence of \textit{HMM} to extract primitives from raw data. \textit{HMM} is a time series model containing two types of Markov processes. Its mathematical form is usually expressed as (1).
$$
\lambda = (\mathcal{A}, \mathcal{B}, \Pi)      \eqno{(1)}
$$

The specific meanings of $\mathcal{A}$,  $\mathcal{B}$ and \textit{$\Pi$} are shown in (2)-(4), where $\mathcal{A}$ represents state transition matrix, $\mathcal{B}$ represents observational state probability matrix and \textit{$\Pi$} is initial probability distribution of hidden state.
$$
\mathcal{A}={[a_{ij}]}_{N \times N} ,a_{ij} = P(i_{t+1} = q_{j} | i_{t} = q_{i})   \eqno{(2)}
$$
$$
\mathcal{B}={[b_{j}(k)]}_{N \times M} ,b_{j}(k) = P(o_{t} = v_{t} | i_{t} = q_{j})   \eqno{(3)}
$$
$$
\Pi={[\pi(i)]}_{N} ,\pi(i) = P(i_{1} = q_{i})   \eqno{(4)}
$$

Here we input 2D laser data as the observation state and regard the environment in which agents are as hidden states. Since laser data is continuous, we choose to use \textit{Gaussian HMM}. The specific meaning of the hidden state represents the distance and intensity of the obstacles. In the actual system, the number of hidden states is set to 10. After training, the laser data can be automatically divided into 10 categories with our \textit{HMM}. Some raw laser data results are displayed in Figure~\ref{fig3}.

Since the classification number of HMM does not have stationary correspondence, we still need to sort the hidden states' number. Here we use the average reward value of each state to sort them, and the definition of the reward function is (5), where \textit{$L_{max}$} represents the max value of laser data, and \textit{l(n)} represents the laser range of \textit{n} degree. 
$$
\mathcal{R}_{collision} = \sum_{n=1}^N[l(n) - L_{max}]    \eqno{(5)}
$$

It can be seen that the number of the reward is related to the area covered by laser, which reflects the complexity around agent.

\subsection{Low-level Control Module}

The left part of Figure~\ref{fig2} is the low-level module, which we call \textit{Control Module}. It contains two independent sub-modules, corresponding to two different sub-tasks.

First, we calculate the speed component of forward and rotating by (6)-(8) with agent position, target position and direction angle of the agent. Therefore, the agent will arrive at the target with the fastest speed and the shortest path without obstacles \cite{27}. In (6), $(X_{t}, Y_{t})$ and $(X_{a}, Y_{a})$ represent the position of target and agent respectively. $yaw_{a}$ is the yaw angle of agent.
$$
desired_{angle} = arctan(\frac{X_{t}-X_{a}}{Y_{t}-Y_{a}}) - yaw_{a}   \eqno{(6)}
$$ 
$$
desired_x = cos(desired_{angle})   \eqno{(7)}
$$
$$
desired_y = sin(desired_{angle})    \eqno{(8)}
$$

In order to complete the task of dynamic collision avoidance, the second sub-module is a \textit{DRL} network, whose state space $\mathcal{S}$ is laser data and action space $\mathcal{A}$ is the velocity component of forward and rotating. Since we only need to evaluate the situation around the current agent, it doesn't have to do with the specific location of the agent. Thus, there is no need to introduce coordinate information into $\mathcal{S}$. This \textit{DRL} network adopts the training method of \textit{DDPG}, and its specific training parameters are shown in Table~\ref{tab1}. 

Compared with previous end-to-end framework \cite{2, 3}, the reward in the training process is more concise. It contains only two parts, one is the coverage of 2D laser in (5), and the other is the punishment after the collision. To ensure that the result of the final output layer of the network is in a reasonable range, the last activation function is \textit{tanh}, so that both $v_x$ and $v_z$ can be restricted to [-1, 1].

\subsection{Action Selection}

After getting the environmental evaluation results and the two kinds of control action of low-level, The fusion of the two actions are needed to get a final control output. We use the weighted average by adjusting the proportion of the outputs of the two sub-modules of low-level according to the re-ordering results of the hidden state of high-level. The specific weighting formulation is shown in (9).
$$
v_{output} = \lambda_1\times v_{target} + \lambda_2 \lambda_s \times v_{collision}  \eqno{(9)}
$$

Based on such a fusion method, the target-driven task will be carried out directly in the case of less surrounding obstacles. In the case of large amount of surrounding obstacles, the priority should be the task of avoiding obstacles. 

\section{EXPERIMENTS}

In this section, we will give a detailed introduction of comparison experiments between previous methods and ours. The simulation environment of these experiments is built with \textit{ROS}\footnote{http://www.ros.org/} and \textit{Gazebo}\footnote{http://www.gazebosim.org/}. The neural network is built with \textit{PyTorch}\footnote{https://pytorch.org/}.

\subsection{Evaluation Criterion}

In these experiments, we choose multiple criteria to make a multi-faceted comparison. These criteria include:
\begin{itemize}
\item the convergence time of model training
\item the success rate of reaching the target
\item the collision rate with other agents
\item the time spent during navigation
\item the trajectory length during the navigation. 
\end{itemize}

The first one is mainly to compare training difficulty among end-to-end \textit{DRL} method, supervised learning method and our method. The second one is the most important criterion. It reflects the ability of the system to completely avoid obstacles and reach the target point. The third one represents the collision times with other agents of all samples. The latter two are auxiliary evaluation criteria, which are used to further explore the efficiency of each algorithm.

\begin{table}[th]
\begin{center}
\begin{threeparttable}
\caption{Parameters of Our system}
\label{tab1}
		\begin{tabular}{c||c}
		\hline
		Parameter Name & Parameter Value\\
		\hline
		Max speed in ORCA & 1\\
		Max neighbors in ORCA & 10 \\
		Neighbor distance in ORCA & 2 \\
		Protect radius in ORCA & 0.4 \\
		Radius in ORCA & 0.1 \\
		Time horizon in ORCA & 0.1 \\
		\hline
		Learning rate of actor & 0.0001\\
		Learning rate of critic & 0.00001\\
		Batch size & 128\\
		$\tau$ in DDPG\tnote{1} & 0.99\\
		$\gamma$ in DDPG\tnote{2} & 0.001\\
		Max step in one episode & 30\\
		Delay of control & 0.1s\\
		\hline
		$\lambda_1$ in (9) & 1\\
		$\lambda_2$ in (9) & 0.4\\
		Hidden state number of HMM & 10\\
		Max laser range & 3.5\\
		Range of $v_x$ and $v_z$ & [-1, 1]\\
		\hline
		\end{tabular}
\begin{tablenotes}
\item[1] $\tau$ is the weight of new model parameter updating.
\item[2] $\gamma$ is the discount of the reward.
\end{tablenotes}
\end{threeparttable}
\end{center}
\end{table}

\subsection{Experiment Settings}

The whole simulation platform is built with \textit{Gazebo}, and the robot model is \textit{TurtleBot3}. \textit{TurtleBot3} is a differential driven robot which has two active wheels and one driven wheel. It has a depth image sensor and a 2D laser, but we only use the 2D laser in this paper. The simulation scene is a square plane, whose size is 12m$\times$12m. In the experiment of 2, 4, 8 and 16 agents, we put the target at the farthest point on the diagonal line of each agent. While in the experiment of 20 and 32 agents, we set the position of agents and targets randomly. The source code of the platform has been released\footnote{https://github.com/GilgameshD/Gazebo-DRL-Navigation}.

Parameters of the \textit{ORCA} method are shown in Table~\ref{tab1}, which have been tuned to fit the experiment scene. Parameters of \textit{DDPG} and \textit{HMM} are also shown in  Table~\ref{tab1}.

\begin{figure}[]
\centering
\includegraphics[width=8.1cm]{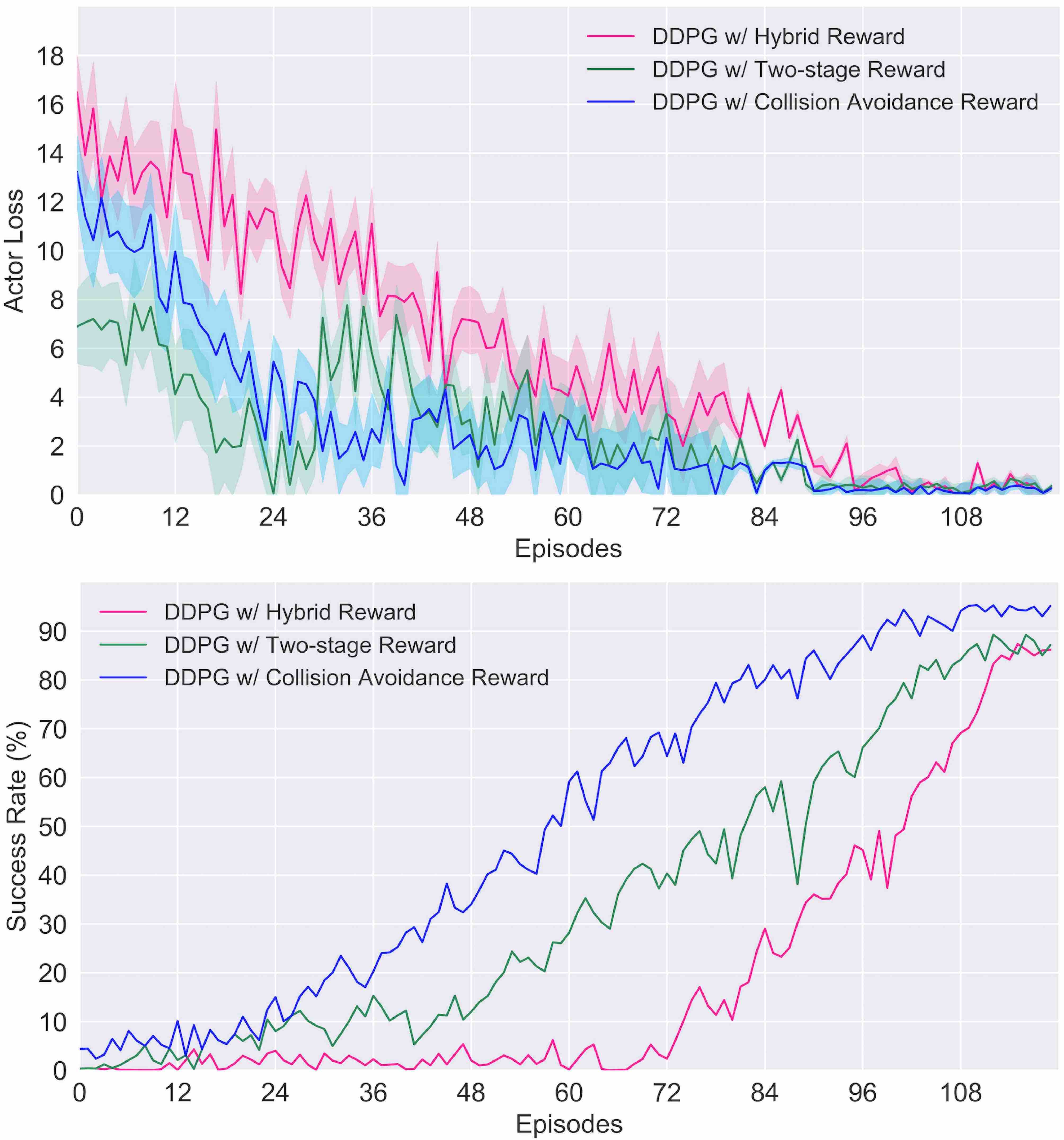}
\caption{\textbf{Actor Loss and Success Rate.} The top figure shows the actor loss of three different methods during training. The bottom figure shows the success rate of three different methods with 20 agents.}
\label{fig4}
\end{figure}

\begin{table*}[th]
\begin{center}
\begin{threeparttable}
\caption{Comprehensive Experiment Result (mean/std)}
\label{tab2}
		\begin{tabular}{c||c||c|c|c|c|c|c}
		\hline
		\multicolumn{2}{c||}{Conditions}&\multicolumn{6}{c}{Agent Number}\\
		\hline
		Metric Name & Method Name & 2 & 4 & 8 & 16 & 20 & 32\\
		\hline
		
		\multirow{4}*{Success Rate}   
		& OCRA\cite{10}                             & {1.000}       & \textbf{1.000}    & 0.982/0.141       & 0.971/0.121     & 0.850/0.211                 & 0.724/0.134\\
		& Supervised Learning                   & {1.000}       &  0.951/0.213    & 0.882/0.141      & 0.722/1.331        & 0.512/2.233                & 0.190/0.344\\
		& Reinforcement Learning\cite{2}   & {1.000}   & \textbf{1.000}     & \textbf{1.000}      & \textbf{1.000}     & 0.867/1.435                & 0.817/1.214\\
		& HNRN                                          & {1.000}   & \textbf{1.000}     & \textbf{1.000}      & \textbf{1.000}     & \textbf{0.941/0.132 }   & \textbf{0.842/1.309}\\
		\hline
		
		\multirow{4}*{Collision Rate} 
		& OCRA\cite{10}                             & {0.000}     & \textbf{0.000}     & \textbf{0.000}     & \textbf{0.000}     & 0.041/0.003               & 0.133/0.014\\
		& Supervised Learning                   & {0.000}     & 0.021/0.123       & 0.102/0.033        & 0.332/0.003       & 0.332/0.003                & 0.490/0.004\\
		& Reinforcement Learning\cite{2}   & {0.000}     & \textbf{0.000}     & \textbf{0.000}     & \textbf{0.000}     & \textbf{0.023/0.145}   & 0.143/0.034\\
		& HNRN                                          & {0.000}     & \textbf{0.000}     & \textbf{0.000}     & \textbf{0.000}     & 0.036/0.001                & \textbf{0.110/0.003}\\
		\hline

		\multirow{4}*{Time Spent\tnote{1} (s)}        
		& OCRA\cite{10}                              & 2.419/0.115     & 5.435/0.208               & 8.354/0.134                 & 15.357/1.207                 & 20.132/3.134                & 31.592/4.524\\
		& Supervised Learning                    & 3.232/0.134     & 6.642/0.133                & 13.731/0.509              & 26.238/2.419                 & 32.346/0.368                & 43.310/6.107\\
		& Reinforcement Learning\cite{2}    & 2.123/0.213     & \textbf{4.145/0.189}    & 6.919/0.435                & \textbf{10.312/1.312}    & 16.145/1.013                & 25.490/3.245\\
		& HNRN                                           & \textbf{2.087/0.233}     & 4.731/0.123    & \textbf{6.242/0.213}    & 11.248/2.426                & \textbf{14.467/2.191}    & \textbf{20.421/4.214}\\
		\hline

		\multirow{4}*{Trajectory Length\tnote{2} (m)}   
		& OCRA\cite{10}                              & 5.231/0.015                 & 8.562/0.345                & 15.978/1.343              & 26.562/1.103               & 37.134/1.303                 & 47.294/2.145\\
		& Supervised Learning                    & 6.125/0.046                 & 11.396/0.506               & 17.432/2.415              & 33.432/2.361              & 40.031/2.034                  & 57.445/3.917\\
		& Reinforcement Learning\cite{2}    & \textbf{5.037/0.023}     & \textbf{6.332/0.113}    & 9.379/0.437                 & 15.334/1.112              & 20.332/0.136                 & 30.198/1.469\\
		& HNRN                                           & 5.128/0.017                 & 7.242/0.233                & \textbf{8.931/0.209}     & \textbf{12.837/1.203}  & \textbf{16.332/0.429}     & \textbf{25.460/1.341}\\
		\hline
		\end{tabular}
\begin{tablenotes}
\item[1,2] Time spent and trajectory length are recorded when the last agent successfully reach its target.
\end{tablenotes}
\end{threeparttable}
\end{center}
\end{table*}

\subsection{Convergence Rate}

In order to compare the convergence rate of different methods, we recorded the training loss of actor in \textit{DDPG}, which is shown in Figure~\ref{fig4}. We display three methods here. The first one is an end-to-end \textit{DRL} framework with a hybrid reward, in which the agent learns target-driven and collision avoidance at the same time. The second one is a two-stage \textit{DRL} method. Its training process is divided into an obstacle-free environment and a collision avoidance environment. We first train the agent in a clear scene with a target, and then change the environment into a multi-agent scene. The third method only has a collision avoidance reward, and it is the model we use in our \textit{HNRN} method.

From the top image of Figure~\ref{fig4}, we can see that the hybrid reward method has a low convergence rate than the other two. Although the two-stage method has a faster convergence at the first stage, it reaches the minimum value later than the third method. Because the two-stage method has a rebound when the second stage begins.

Besides the actor loss, we also present the success rate of 20 agents. According to the bottom image of Figure~\ref{fig4}, the first method (Red line) has a near-zero success rate during the first-half phase of training. This phenomenon can be explained by the fact that the hybrid reward is much difficult to train. The agent is usually confused by the complex reward function. Our method (Blue line) achieve some success case at the beginning of the training because of its target-driven policy. And after several epochs, our method achieves a better result than the other two methods. 

\subsection{Comprehensive Experiment}

Some comprehensive experiments are designed and implemented to evaluate the performance of our method. The result of these experiments is displayed in Table~\ref{tab2}. We select four methods to compare， and we implement the experiment on several numbers of agent to study the influence of the density of the agent. As the agent number increases, the spent time and trajectory length increase. The reason is that the more agents there are, the less free space they have. Thus, they find it more difficult to search a safe path to the target.

Reinforcement learning method \cite{2} and our \textit{HNRN} method behave almost the same in a small agent number environment. However, in the scene of 20 and 32 agents, our method achieves a better result than DRL method \cite{2}. This result proves that our method performs better in a complicated environment because of the hierarchical structure. As for the success rate, our method also outperforms other methods. The collision rate represents the situation that one agent hits another one.  It can be concluded from the Table~\ref{tab2} that collision scenes account for most of the failure cases in out method, rather than trapped by other agents. In contrast, most of the failure cases of \cite{2} are trapped cases. The reason is that the agent trained by hybrid reward doesn't learn enough strategy to deal with much complicated environment.

Traditional method \textit{OCRA} \cite{10} successes in most cases, both simple and complicated. However, this method spends more time and longer trajectory to get to the target in the complicated scene. The lack of learning-based policy accounts for this result.

The performance of the supervised learning method is the worst, whose training data comes from the traditional method \textit{OCRA} \cite{10}. The fact that the amount of training data is not enough to cover all situations limits this method. For some simple environment, the agent trained by supervised learning can reach the target, but in a more complicated scene, it fails.

\section{CONCLUSIONS}

In this paper, we propose a hierarchical navigation framework, which utilizes reinforcement learning and \textit{HMM}. Our method consists of two levels. In high-level architecture, an \textit{HMM} is trained to evaluate the agent's environment in order to obtain a score. According to this score, adaptive control action will be chosen. In low-level architecture, two sub-systems are introduced, one is a differential target-driven system, which aims at heading to the target, the other is a collision avoidance \textit{DRL} system, which is used for avoiding obstacles in the multi-agent environment. We compared our method with traditional \textit{ORCA} method, supervised learning method and a two-stage reinforcement learning method. The experiment results demonstrate that ours outperform others in the complicated environment.

There are still some limitations of this work. We only evaluate our algorithm in simulation platform, and the navigation scene is quite simple. Some static obstacles and scenes of complex shape should be tested for more detailed verification. The shape of the robot is also a significant factor for navigation algorithm, thus heterogeneous shapes should be included in future work. In spite of all these insufficient conditions, our work briefly proves the advantages of a hierarchical framework for target-driven tasks.

\bibliographystyle{IEEEtran}
\bibliography{mybib}

\end{document}